\documentclass[a4paper,             
               13pt,                
               twoside,             
               numbers=noenddot,    
               headinclude,         
               footexclude,         
               headsepline,         
               smallheadings,       
               openany,             
               cleardoubleempty,    
               tablecaptionabove,   
              parskip=half,         
              plainheadsepline,     
              toc=left              
               ]{scrbook}           
\usepackage[english]{babel}         
\usepackage{graphicx}               
\usepackage[caption=false]{subfig}  
\usepackage[intlimits]{amsmath}     
\usepackage{amssymb,amsfonts}
\usepackage{mathpazo}               
\usepackage[manualmark]{scrpage2}   
\usepackage[
            pdftitle={Conference Autonomous Systems},
            pdfauthor={Herwig Unger, Wolfgang A. Halang (Eds.)},
            pdfsubject={},
            pdfkeywords={Proceedings, Autonomous Systems, Conference},
            pdfborder={0 0 0},     
            colorlinks=false,      
            breaklinks             
            ]{hyperref}            
\usepackage[sort&compress,square,comma,numbers]{natbib}
\usepackage{hypernat}              
\usepackage[sectionbib]{chapterbib}

\usepackage{multirow}
\usepackage{url}
\usepackage{wrapfig}

\usepackage{titlesec}
\titlespacing{\section}{0pt}{6pt}{6pt}
\titlespacing{\subsection}{0pt}{4pt}{4pt}
\titlespacing{\subsubsection}{0pt}{4pt}{4pt}
\titlespacing{\paragraph}{0pt}{4pt}{4pt}

\makeatletter
\renewcommand*\l@chapter{\@dottedtocline{0}{3em}{0em}}
\makeatother

\newcommand*{\achapter}[3][\chapteroptarg]{%
  \def\chapteroptarg{#2}%
  \addchap[\sffamily\bfseries#2\protect\newline{\mdseries#3}\vspace*{1em}]
    {#2}
  \markboth{#3}{#1}
}
\clearscrheadfoot  
\ihead{\headmark}  
\ohead[\pagemark]{\pagemark}  
\typearea[18mm]{18}                 %

\addtokomafont{caption}{\sffamily} 
\setkomafont{captionlabel}{\sffamily\bfseries}
\addtokomafont{chapter}{\centering} 

\addto\captionsenglish{}

\begin{document}
\def\figurename{Fig.}

\frontmatter
\pagenumbering{roman}              


\cleardoublepage                   
\setcounter{page}{5}
\setcounter{tocdepth}{0}

\cleardoublepage


\mainmatter

\achapter{Towards Unsupervised Familiar Scene Recognition in Egocentric Videos}{E. Talavera, N. Petkov and P. Radeva}

\begin{center}
Estefan\'ia Talavera$^{1,2}$, Nicolai Petkov$^2$ and Petia Radeva$^{1,3}$
\par

$^1$University of Groningen, Netherlands, $^2$University of Barcelona, Spain, $^3$Computer Vision Center, Spain
\end{center}

\begin{quote}
\emph{Abstract:}

Nowadays, there is an upsurge of interest in using lifelogging devices. Such devices generate huge amounts of image data; consequently, the need for automatic methods for analyzing and summarizing these data is drastically increasing. We present a new method for familiar scene recognition in egocentric videos, based on background pattern detection through automatically configurable COSFIRE filters. We present some experiments over egocentric data acquired with the Narrative Clip.

\end{quote}

\section{Introduction}

\begin{wrapfigure}{r}{0.45\textwidth}
  \begin{center}
  \vspace{-1em}
    \includegraphics[width=0.4\textwidth,height=0.36\textwidth]{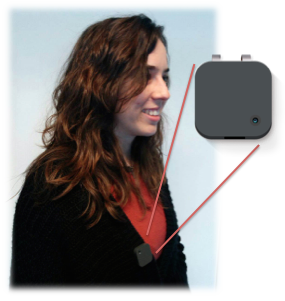}
  \end{center}
\caption{Wearable camera - Narrative Clip}
\vspace{-1.5em}
\label{talavera:cameras}
\end{wrapfigure}

Lifelogging, described as the digital capture of a person's everyday activities by wearing a digital recording device, is recently showing its potential as a new tool for user life-style analysis, with the aim of improving life-habits. By recording the person's own view of the world, lifelogging opens new questions and leads to the desired and personalized analysis of the camera wearer's lifestyle. Egocentric videos are usually automatically recorded by wearable cameras (see Fig. \ref{talavera:cameras}). These videos offer a first-person view of the world, describing daily activities and providing information about the interaction of the user with the surrounding environment. This huge collection of data is considered as a tool by behavioral specialist such as psychologists since they can describe user's activity patterns. Specialists and users are both interested in gaining an overview and finding important events of interest. Some of the possible applications could be the analysis and summarization of the lifelogs data into narratives, offering a memory aid to mild cognitive impairment (MCI) patients by reactivating their memory capabilities \cite{hodges2006sensecam}, or to provide activity diary based on what, where and with whom the user performs any possible activity, related to health or not. 

\vspace{-1em}

Different other approaches from the literature focus on the detection and tracking of the people surrounding the user, the objects they manipulate or the food they eat (e.g. \cite{BOT,fathi2011learning,bolanos2013active}, respectively) from the life-logging videos since they usually represent the information that users want to retrieve. The approach presented in reference \cite{expstudy} introduced the visual lifelogs analysis as a tool to trigger autobiographical memory about past events. 

\begin{figure}[ht!]
\begin{center}
\includegraphics[width=1\linewidth,height=0.35\linewidth]{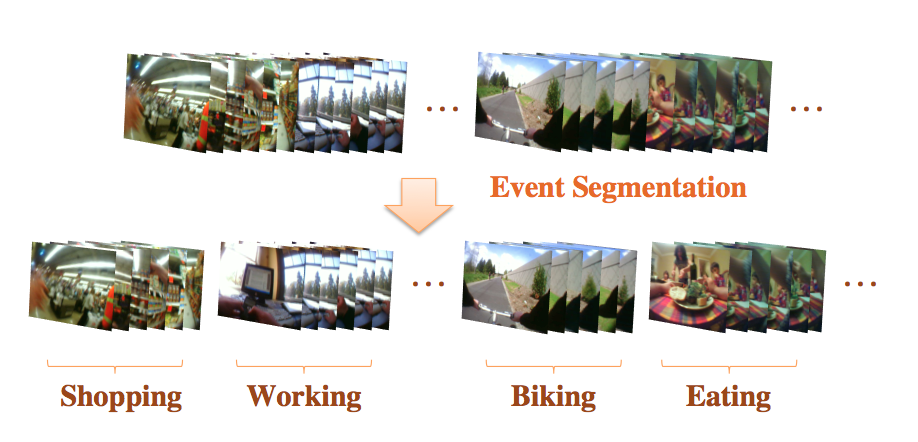}
\caption{Example of temporal segmentation}
\label{talavera:segmentation}
\end{center}
\end{figure}

Video segmentation aims to split the video into different groups of consecutive images called \textit{events} or \textit{scenes} that describe the performance of an activity or a specific environment where the user is spending time (see Figure \ref{talavera:segmentation}). Many segmentation techniques have been proposed in the literature in an attempt to deal with this problem, such as video summarization based on clustering methods or on object detection. The work described in \cite{userkeyp} was a first approach where the user selected the frames considered important as key frame (considered as the frame that best represents the scene), generating the storyboard that reported object's trajectory. Other studies incorporate audio or linguistic information \cite{audio,lengunder} to the segmentation approach looking for the semantic meaning of the video. 
\vspace{-2em}

\vspace{-3em}

A suitable video segmentation with its later summarization, by selecting the images that best describe the segments, aims to give a better overview than the daily set of photos, which can be composed of thousands of frames. Therefore, we propose a method that goes further than the works presented above, and than our previous work \cite{talavera2015r-clustering} in which we focus on the temporal egocentric data segmentation. This method seeks the recognition of important events within the segmented data. We aim to recognize \textit{where} the user is. From that, we aim to infer \textit{what} he/she is doing and \textit{for how long}.

Lifelogging images can be very different despite having been recorded during the performance of the same activity, due to the constant user's movement that leads into a changing and challenging background. However, some characteristic patterns of the scene often remain present in these event frames. With this in mind, we want to detect the user's familiar scenes, which we associate with daily activities performed by the user, such as spending time at the lunch room, working in front of the pc, driving, etc. In contrast to the majority of the works presented in the literature, which focus on the analysis of videos recorded with a high temporal resolution such as 20 or 30 fps, we work with a wearable device with a low temporal resolution such as 2-3 fpm with changing background and scenes during the video. 

In this paper, on one hand, we apply a video segmentation step, using a method introduced in \cite{talavera2015r-clustering}, where the video is divided into different events based on low-level features of the frames. We use Convolutional Neural Network (CNN) vector activation over the entire image as a global image feature descriptor. The CNN features, which are able to focus just in the environment appearance, are normalized and compared in order to define the affinity value with the representative image of the frame. On the other hand, we introduce a \textbf{scene identification approach} which bases the labelling of the images on the responses of the pre-trained COSFIRE filters, which are configured using specific patterns from sample images of the environment of the scene. Based on the excellent accuracy achieved by these filters, we based the scene detection on their responses. We illustrate the proposed method for one familiar event recognition from our dataset. In the next section, we detail the proposed approach. In section \ref{talavera:results}, we discuss experimental results and, finally,  in section \ref{talavera:conclusions} we draw some conclusions.

\section{Familiar Scene Recognition Approach}

Given an egocentric sequence of images, our goal is the recognition of some usual/familiar environment. By discovering this information we may be able to infer when and for how long the user performs a specific activity. Therefore, given the problem of scene identification in egocentric sets of data, our contribution is a method for scene's background detection within segmented egocentric data that relies on trainable COSFIRE filters. 

\vspace{-0.5em}

\subsection{R-Clustering: Temporal Video Segmentation}

Given the problem of temporal segmentation of egocentric videos, we apply R-Clustering, introduced in reference \cite{talavera2015r-clustering}. The clustering technique is applied over the image features, extracted from the convolutional neural network trained on ImageNet \cite{NIPS2012_4824}.
\vspace{-2em}
\begin{figure}[ht!]
\begin{center}
\includegraphics[width=1\linewidth]{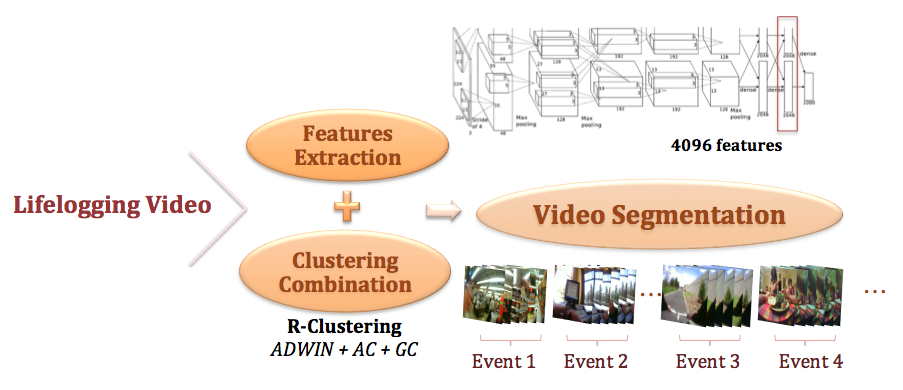}
\caption{R-Clustering Model}
\label{talavera:R-Clustering}
\end{center}
\end{figure}
\vspace{-5em}

\begin{figure}[ht!]
    \vspace{-1em}
    \centering
    \includegraphics[width=1.0\linewidth]{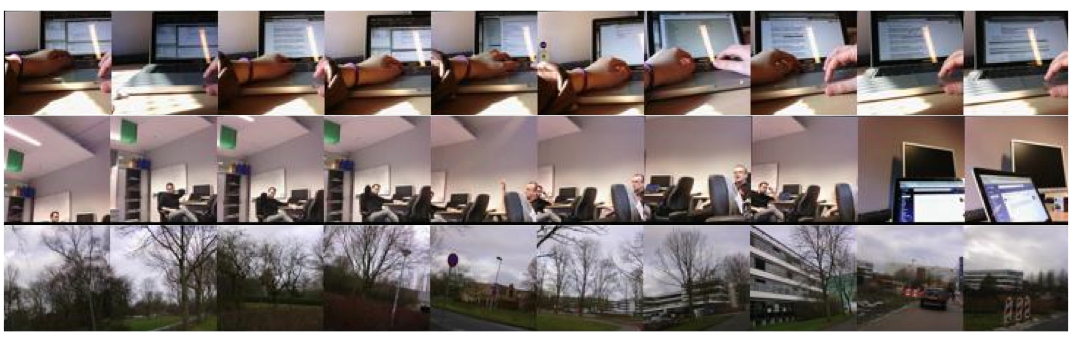}
    \caption{Illustration of R-Clustering segmentation results for 3 events from a Narrative set}
    \label{talavera:segmentation_results2}
 \vspace{-2em}
\end{figure}

\subsection{Scene recognition} 

\subsubsection{Trainable COSFIRE Filters}
 
The Combination Of Shifted FIlter REsponses (COSFIRE) trainable filter was first introduced in \cite{COSFIREmethod}. It is a key point detector with a wide range of possible applications in computer vision. It was inspired by and shares similar properties with shape-selective neurons in visual cortex, multiplying the responses of sub units that are sensitive to specific curve patterns. The COSFIRE filter, given a prototype image and a key point of a region of interest, is automatically configured from this region of interests and can be used as pattern detector over other images. For the automatic configuration of the filter a training pattern is presented and certain blur and shift parameter values are determined and channels of a bank of Gabor selected. The resulting COSFIRE filter is tolerant to rotation variations and to slight deformations.The filter's response is computed as the weighted geometric mean of the shifted and blurred responses of the selected Gabor filters. Therefore, COSFIRE filters give a response only when all constituent parts of the pattern of interest are present. 
\vspace{-2.5em}
\begin{figure}[ht!]
\includegraphics[width=\linewidth,height=0.4\linewidth]{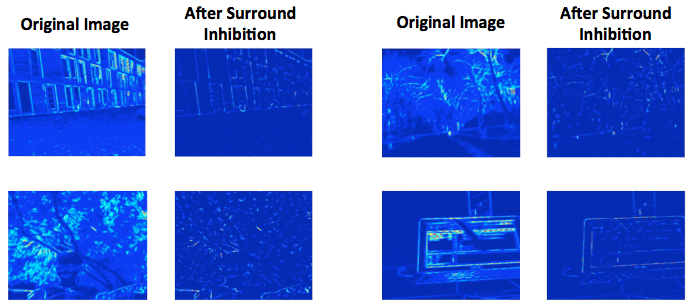}
\caption{Response images of the edge detector. Enhancement of occluding contours and region boundaries by suppression of texture edges}
\label{talavera:inhibition}
\end{figure}
\vspace{-2em}

\textbf{Dealing with texture noise} 

The original COSFIRE filters as proposed in \cite{COSFIREmethod} give many false responses in areas of texture. Therefore, the output will not be very useful for contour based object recognition since it will respond to wrong areas due to the high probably of finding any pattern within an area rich in texture. We face this problem by modifying the results of the Gabor-energy filtering to enhance the region boundaries by the suppression of texture edges, i.e. we apply surrounding inhibition (see references \cite{Nicolai1} and \cite{Nicolai2}). The texture edges disappear from the output image that is now representing the main structural information (see Fig. \ref{talavera:inhibition}). This is more useful for contour based object recognition and helps for the detection of objects of interest in rich texture areas. 
\vspace{-0.5em}

\textbf{COSFIRE filters application}
\vspace{-0.5em}

Given an egocentric video the method proposed labels the images with a scene tag or the 'unknown' tag. The label decision is based on the responses obtained by applying COSFIRE filters to the video frames. Therefore, for each input image all filters are applied. If an image contains the pattern with which a given filter was configured there will be a significant response by that filter. The images receive a scene label if the filters give a response due to the presence of the pattern. If several filters respond to an image, we assign the label of the scene from which there have more filters responses. If no filter responds to an image, it will be labelled as 'unknown' scene.
\vspace{-2.5em}
\begin{figure}[ht!]
\includegraphics[width=\linewidth]{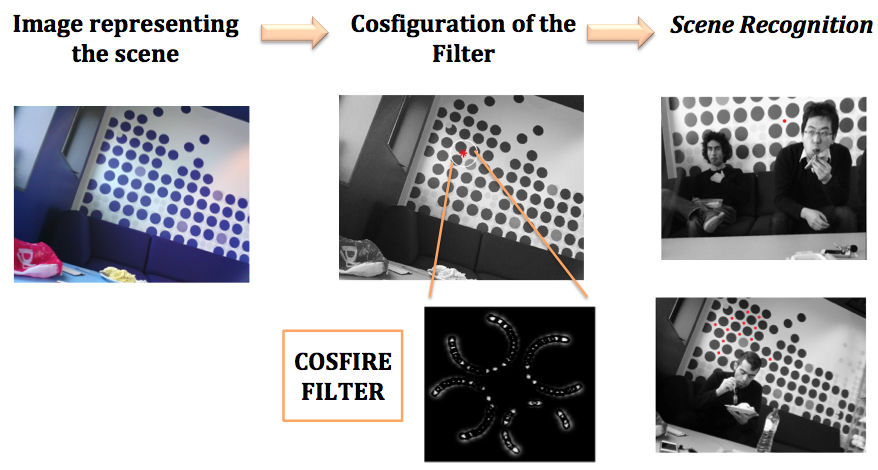}
\caption{Patterns of interest for a sample image. The COSFIRE filter is configured based on the pattern indicated by the highlighted keypoint and its surroundings shown in the upper middle panel. The two right images show how the pattern has been discovered over test images by applying the trained COSFIRE filter.}
\label{talavera:CosfireConfiguration}
\end{figure}
\vspace{-2.5em}

\section{Experiments} 
\label{talavera:results}

In this section, we present the dataset on which we test our method, that has been recorded by us, and the statistical validation measurements that we are going to use in order to evaluate the results. We show an example of trained filters applied to egocentric data for scene recognition.
 
	\subsection{Data Acquisition and pre-Processing}
\begin{wrapfigure}{r}{0.45\textwidth}
  \begin{center}
  \vspace{-4em}
\includegraphics[width=0.4\textwidth]{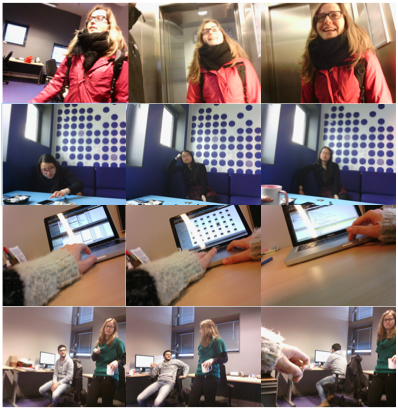}
  \end{center}
  \vspace{-1.5em}
\caption{Example of Narrative frames} 
\vspace{-3em}
\label{talavera:imagescamera}
\end{wrapfigure}
Our dataset is composed by egocentric videos recording the user's daily indoor/outdoor activities (working with a laptop, walking, eating, etc). These videos were recorded with a Narrative Clip camera (http://getnarrative.com/), with a resolution of 2fpm and a normal lens. The user wore the wearable camera fixed to his/her chest, so the frames vary following the user's movement. These movements and the wide range of different situations that the user experiences during his/her day leads to new challenges such as background scene variation, changes in lighting conditions, and handled objects appearing and disappearing during the video.

The dataset we use to test our method contains 787 images with a mix of indoor and outdoor scenes. Since we defined the patterns of interest from where the filters where configured, we have manually created a ground-truth for which we define the images from the dataset that present the patterns describing the scenes we are looking for.

	\subsection{Evaluation Criteria} 
To evaluate the effectiveness of the scene detection approach we use the F-Measure (FM) (following \cite{li2013daily}): $FM= 2(RP)/(R+P)$, where $P$ is the precision $(P=TP/(TP+FP)$, $R$ is the recall $(R=TP/(TP+FN)$ and $TP$, $FP$ and $FN$ respectively are the number of true positives, false positives and false negatives of the video scene's frames detected. The true positives correspond to the images where the method detects the pattern and it is present, and the false positives are the images with responses to the filters but that are not representing the scene and/or the pattern from which these filters where configured from. The false negatives represent the images that have the pattern but were not detected by the filters and were not labelled as belonging to any of the activity, therefore labelled as 'unknown'. Lower values of the F-Measure represent the result of a bad scene detection while higher values come from a good recognition, with a maximum value 1 where the images labelled are the images labelled by the user as presenting the pattern of interest.
\vspace{-2.5em}
\begin{figure}[ht!]
\includegraphics[width=\linewidth]{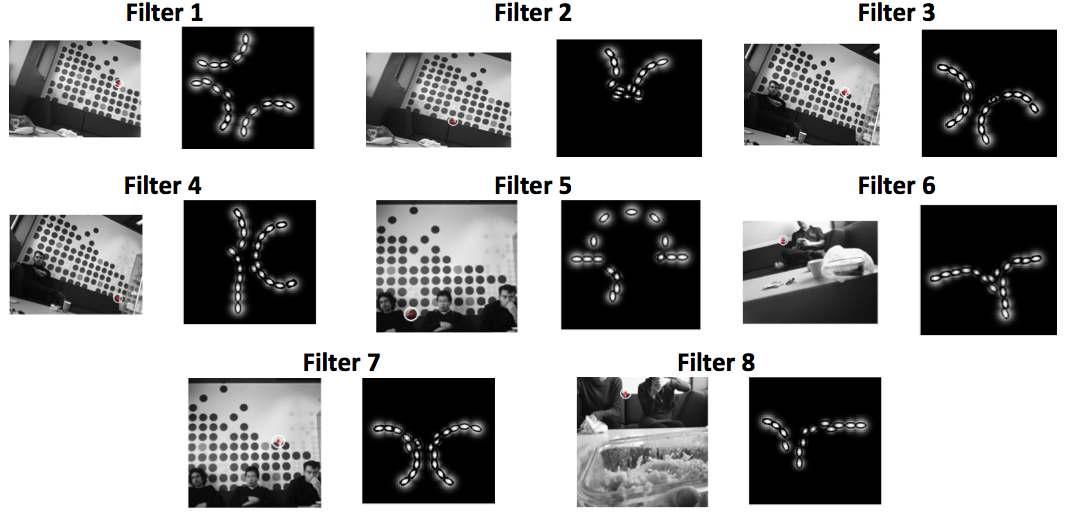}
\caption{Patterns of interest for the CoffeeCorner scene and the structure of the corresponding COSFIRE filters}
\vspace{-1em}
\label{talavera:configCosf_CC}
\end{figure}
\vspace{-3.5em}
\begin{figure}[ht!]
\includegraphics[width=\linewidth]{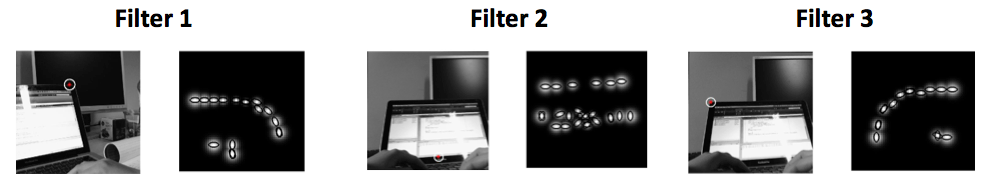}
\caption{Patterns of interest for the Working scene}
\vspace{-1em}
\label{talavera:configCosf_Lp}
\end{figure}
\vspace{-2em}

\subsection{Examples of the performance of the method} 

For this work and as a probe of concept, we focus on the discovery of two familiar scenes within the wide activities that can be performed during the day. These scenes are 'CoffeeCorner' or 'Working'. As described before, for each of these scenes we focus on characteristic patterns or objects that describe the environment of the scene. For the 'Working' activity we look for the presence of a computer in the images, while for the 'CoffeeCorner' scene we look for different patterns describing the background of the room where the user is having lunch and resting. For each of the key-points describing the scenes we have some sample images. We train 8 (CoffeeCorner - see Fig \ref{talavera:configCosf_CC} ) and 3 (Working) filters per scene, from 5 and 2 sample images respectively. We apply them to the test dataset and obtain the following results - shown in Table 1.

\begin{table}[ht!]
\vspace*{-1.5em}
\centering
\begin{tabular}{c||c|c|c|c|c|c}
 Scenes & TP &FN&FP&Precision & Recall & F-Measure \\ \hline
CoffeeCorner  & 89 & 10 & 5 & 0.95 & 0.89 & \textbf{0.92}  \\
Working & 107 & 26 & 107 & 0.50 & 0.80 & \textbf{0.62}  \\
\end{tabular}
\vspace{1em}
\caption{F-Measure for the tested method over our egocentric datasets}
\label{tab:results}
\end{table}
\vspace{-2.5em}

As we show in Table \ref{tab:results}, for the recognition of the CoffeeCorner scene we obtain better results. The trained filters for this scene (see Fig. \ref{talavera:configCosf_CC}) are more specific than the trained filters for the Working scene (see Fig. \ref{talavera:configCosf_Lp}), which are patterns describing corners that can be found in many other environment. This could be the reason of the worse performance of the second group of filters, their not totally characterization of a singular scene. 

\textbf{The time dedicated to performing an activity} is also important for the user's behavior analysis. The images we are working with provide meta-data with information about the time when they were taken. Therefore, by recognizing present background patterns in these images, we can group them, label them as an event and compute for how long the user was at the environment.
\vspace{-2em}
\begin{figure}[ht!]
\centering
\includegraphics[width=0.8\linewidth,height=0.2\linewidth]{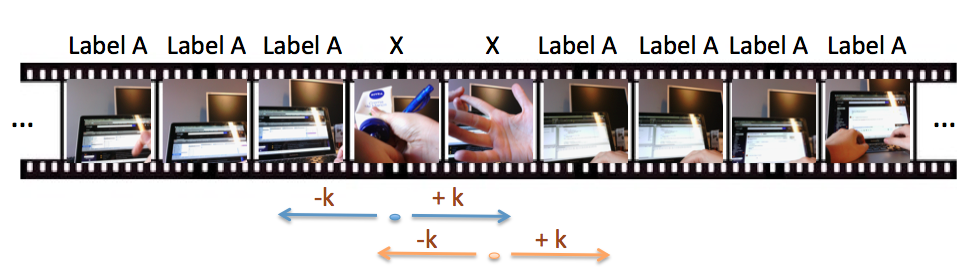}
\caption{Sliding decision window for detected holes within the activities}
\vspace{-1em}
\label{talavera:countingtime}
\end{figure}
\vspace{-1.5em}

We configure filters for some and not all the possible patterns describing a scenario, therefore, the method cannot always label all the images belonging to a scene. This lack of labelling leads into temporal holes within the images labelled as belonging to an activity. It can be due to the occlusion of the searched pattern by an object or a person, or by the constant movement of the user that could leave the scene for a short period of time, etc. To solve this problem, we apply a temporal sliding window to the already labelled images. The idea is that, by looking at the previous and following images from a 'hole image' ( \textit{Im}, image not labelled within the activity but surrounded by labelled images) within the scene, we can decide if \textit{Im} will be label as part of the scene or not basing the decision on the labels assigned to the adjacent images. We call \textit{k} the number of images we consider for this decision and define it as 2.

\vspace{-2em}
\begin{figure}[ht!]
\centering
\includegraphics[width=0.8\linewidth,height=0.45\linewidth]{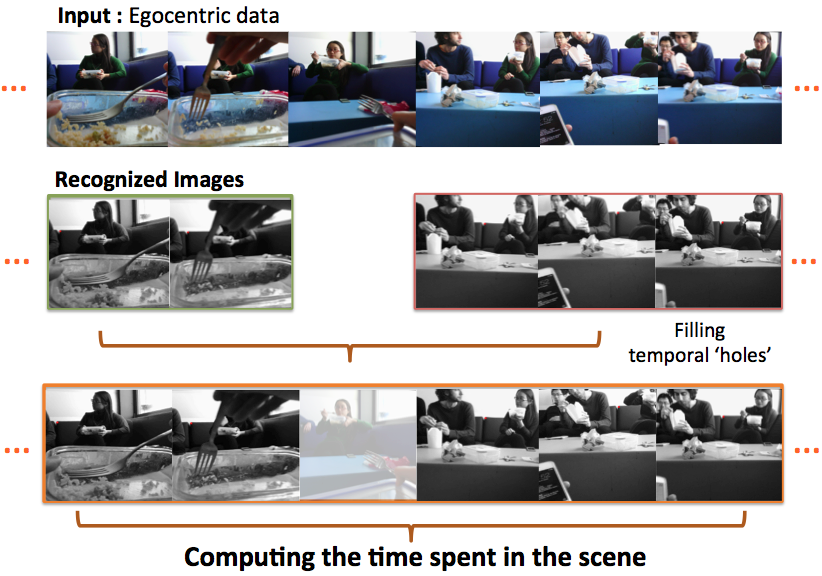}
\vspace{-1em}
\caption{Illustration of the suppression of temporal holes within the images labelled}
\end{figure}
\vspace{-2em}

\section{Conclusion}
\label{talavera:conclusions}

In this work we have presented a novel scene recognition method for egocentric datasets, aiming to identify the time spent performing different activities. Egocentric videos contain frames with changing foreground information but often with a semi-static background. We propose to apply COSFIRE filters for the scene background recognition, since they can be automatically pre-configured from sample images of the scenes. We have applied the method proposed over an egocentric dataset recorded with the Narrative Clip. As a result, we have achieved a F-Measure of 0.78\%. From these results we can conclude that, although the trainable COSFIRE filters are very robust for the recognition of a patterns, for the scene recognition we should select not just characteristic areas but also singular and unique patterns able to distinguish the scene they represent from the rest of the environments. Our approach is a good first attempt for the definition of a new scene recognition model. As future line, we intend to define an automatic event recognition model for video summarization and to extend it for the recognition of a wider range of scenes.

\bibliographystyle{abbrv}

\backmatter

\end{document}